\documentclass{article} 
\usepackage{nips12submit_e,times}
\usepackage{epsfig}
\usepackage{dfg}

\def\figref#1{Figure~\ref{fig:#1}}

\def\figlabel#1{\label{fig:#1}\label{p:#1}}

\newcommand{\enotesoff}{\long\gdef\enote##1##2{}}
\newcommand{\enoteson}{\long\gdef\enote##1##2{{\bf
       \hspace{1cm}$<<<$ ##1: ##2 $>>>$\hspace{1cm}}}}

\enoteson
\enotesoff

\title{Two SVDs produce more focal deep learning representations}

\author{
Hinrich Sch\"{u}tze\\
Center for Information and Language Processing\\
University of Munich, Germany\\
\texttt{hs999@ifnlp.org} \\
\And
Christian Scheible \\
Institute for NLP \\
University of Stuttgart, Germany \\
\texttt{scheibcn@ims.uni-stuttgart.de} 
}

\nipsfinalcopy 

\begin{document}


\maketitle

\begin{abstract}
A key characteristic of work on deep learning and neural
networks in general is that it relies on representations of
the input that support generalization, robust inference,
domain adaptation and other desirable
functionalities. Much recent progress in the
field has focused on efficient and effective methods for
computing representations. In this paper, we propose
an alternative method that is more efficient than prior
work and produces representations
that have a property we call \emph{focality} -- a property
we hypothesize to be important for
neural network representations. The method consists of a
simple application of two consecutive SVDs and is inspired
by \cite{anandkumar12two}.
\end{abstract}

In this paper, we propose to generate representations for
deep learning by two consecutive applications of singular
value decomposition (SVD).  In a setup inspired by
\cite{anandkumar12two}, the first SVD is intended for
denoising. The second SVD rotates the representation to
increase what we call \emph{focality}.  In this initial study, we do not
evaluate the representations in an application. Instead we
employ diagnostic measures that may be useful
in their own right to evaluate the quality of representations
independent of an application.

We use the following terminology.  SVD$^1$ (resp.\ SVD$^2$)
refers to the method using one (resp.\ two) applications of
SVD; 1LAYER (resp.\ 2LAYER) corresponds to a
single-hidden-layer (resp.\ two-hidden-layer) architecture.

In Section 1, we introduce the two methods SVD$^1$ and
SVD$^2$ and show that SVD$^2$ generates better (in a sense
to be defined below) representations than
SVD$^1$. In Section 2, we compare 1LAYER and 2LAYER SVD$^2$
representations and show that 2LAYER representations are
better. Section 3 discusses the results. We present our
conclusions in Section 4.

\section{SVD$^1$ vs.\ SVD$^2$}
Given a base representation of $n$ objects in ${\cal R}^d$,
we first compute the first $k$ dimensions of an SVD on the corresponding $n \times d$
matrix $C$. $C_k=USV^T$ (where $C_k$ is the rank-$k$
approximation of $C$). We then use $US$ to represent each object as a
$k$-dimensional vector. Each vector is normalized to unit
length because our representations are count vectors where
the absolute magnitude of a count contains little useful
information -- what is important is the relative differences
between the counts of different dimensions.
This is the representation
SVD$^1$. It is motivated by standard arguments for
representations produced by dimensionality reduction:
compactness and noise reduction. Denoising is also the
motivation for the first SVD in the method proposed by
\newcite{anandkumar12two}. 


We then perform a second SVD
on the resulting matrix $C'$ of 
dimensionality $ n \times k$.
$C'=U'S'V'^T$ (full-rank, no dimensionality reduction). 
We again use $U'S'$ to represent each object as a
$k$-dimensional vector. Each vector is normalized to unit length.
This is the representation
SVD$^2$.

Note that 
the operation we are applying is not equivalent to a single
linear operation because of the lenght normalization that we
perform between the two SVDs.

SVD$^2$ is
intended to be a rotation of SVD$^1$ that
is
more ``focal''  in the following sense.
Consider a classification problem $f$ over a $k$-dimensional
representation space $R$.
Let $M_\theta(f,R)$ be  the size $k'$ of
the smallest subset of the dimensions that support an
accuracy above a threshold $\theta$ for $f$. Then a representation
$R$ is more focal than $R'$ if $M_\theta(f,R)<M_\theta(f,R')$. 
The intuition is that good deep learning representations
have semantically interpretable hidden units that contribute input
to a decision that is to some extent independent of other
hidden units. We want the second SVD  to rotate the
representation into a ``more focal'' direction.

The role of the second SVD is somewhat analogous to that of
the second SVD in the approach of
\newcite{anandkumar12two}, where the goal also is  to find a
representation that reveals the underlying structure of the
data set.

\begin{figure}[tp]
\includegraphics[width=7cm]{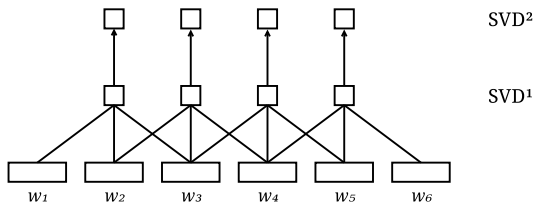}
    \caption{1LAYER  architecture. Distributional word
      vectors form the bottom layer. A trigram --
      represented as a triple of word vectors -- is
      transformed by the first SVD into a 100-dimensional
      vector (layer ``SVD$^1$''). These vectors are then
      rotated (layer ``SVD$^2$'').}
\figlabel{layer1}
\end{figure}

The architecture of the 1LAYER setup is depicted in \figref{layer1}.

{\bf Experimental setup.} We use 
a corpus of movie review sentences
 \cite{Pang+Lee:04a}. Following
\newcite{schutze95distributional}, we first compute a left
vector and a right vector for each word. The dimensionality
of the vectors is 250. Entry $i$ for the left (right) vector of word
$w$ is the number of times that the word with frequency rank
$i$ occurred immediately to the left (right) of $w$. 
Vectors are then tf-idf weighted and length-normalized.
We 
randomly select 100,000 unique trigrams from the corpus,
e.g., ``tangled feelings of'' or ``as it pushes''. Each
trigram is represented as the concatentation of six vectors,
the left and right vectors of the three words. This defines
a matrix of dimensionality $n \times d$ 
($n=100000$,  $d=1500$).  We then compute SVD$^1$ and
SVD$^2$ on this matrix for $k=100$.

\begin{figure}[tp]
\includegraphics[width=5cm]{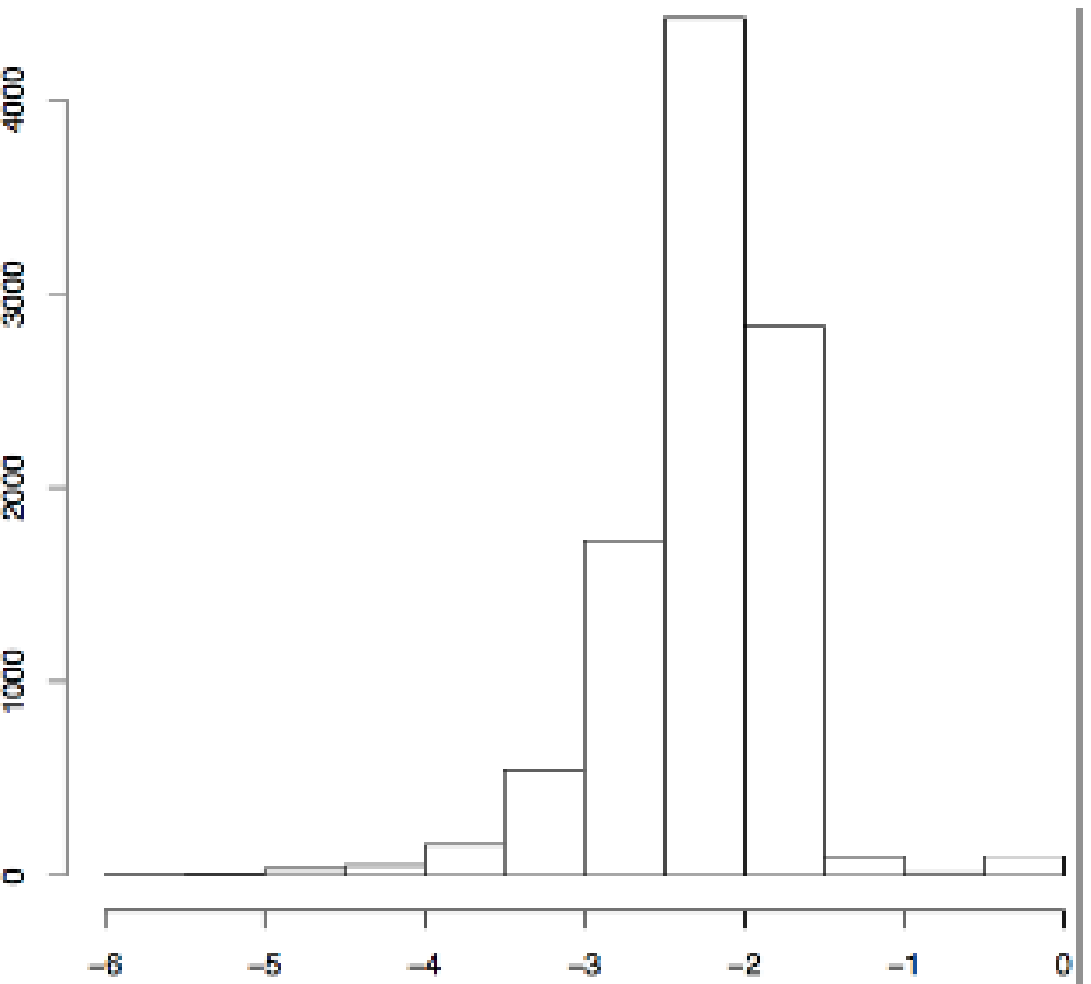}

\includegraphics[width=5cm]{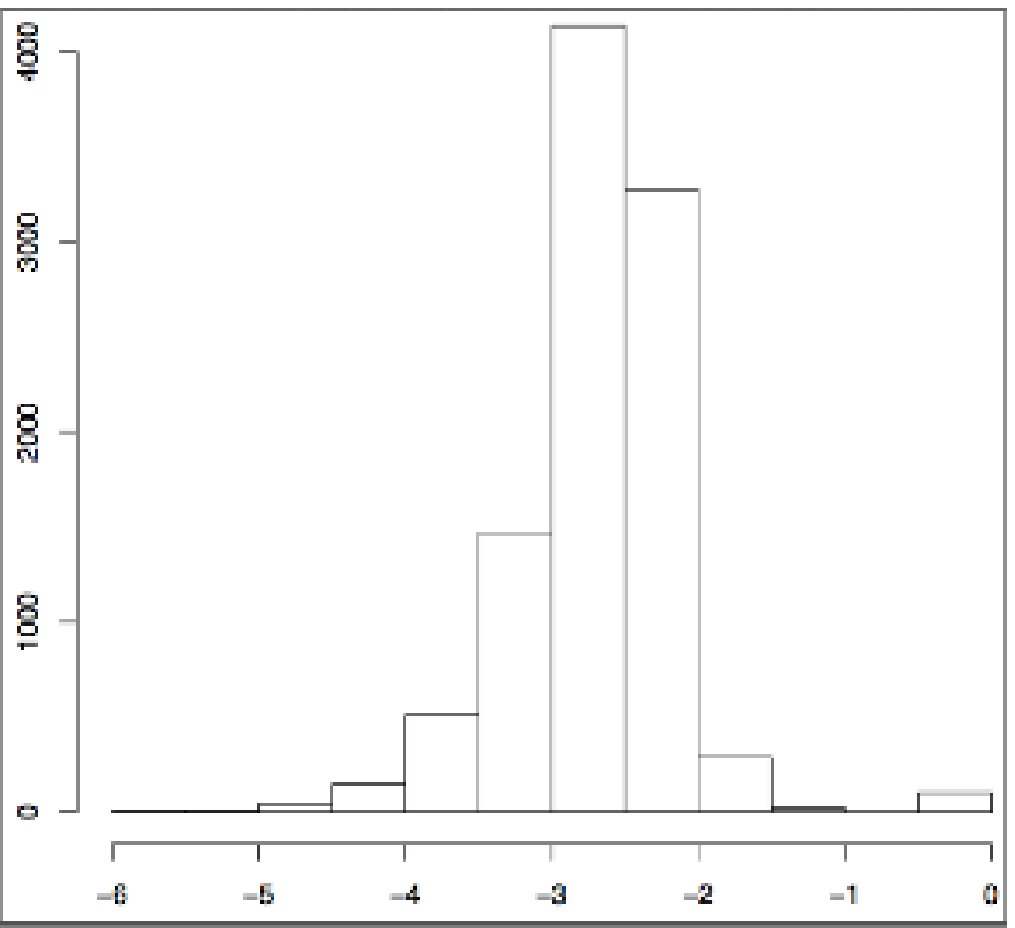}
    \caption{Histograms of $\log_{10}|c|$ of 
the 10,000
      correlation coefficients of SVD$^1$
      (above) and SVD$^2$ (below).}
\figlabel{hist}
\end{figure}

{\bf Analysis of correlation coefficients.}
\figref{hist} shows histograms of the 10,000 correlation
coefficients of SVD$^1$
      (left) and SVD$^2$ (right). Each correlation coefficient
is the correlation of two columns in the corresponding 
100000 $\times$ 100 matrix  and is
transformed using the function $f(c) = \log_{10}|c|$ to
produce a histogram useful for the analysis.
The histogram of SVD$^2$ is shifted by about 0.5 to the
left. This is a factor of $10^{0.5} \approx 3$. Thus, SVD$^2$
dimensions have  correlations that are only a third as large
as SVD$^1$ correlations on average.

We take
this to indicate that SVD$^2$ representations are more focal
than SVD$^1$ representations
because the distribution of correlation  coefficients
would change the way it changes from
SVD$^2$ to
SVD$^1$ if we took a focal representation (in the most
extreme case one where each dimension by itself supported a
decision) and rotated it.

{\bf Discrimination task.}
We randomly selected 200 words in the frequency range
$[25,250]$ from the corpus; and  randomly arranged them into
100 pairs. 
An example of such a pair is (documentary, special).
For each pair, we first
retrieved the SVD$^1$ and SVD$^2$ representations of all triples from the set of 100,000 in which one
of the two words was the central word. 
For the example, 
``typical documentary footage'',
``good documentary can'', and
``really special walk'' are such triples.
Then we determined
for each dimension $i$ of the 100 dimensions (for both SVD$^1$ and SVD$^2$) the
optimal discrimination value $\theta$ by exhaustive search; that is,
we determined the threshold $\theta$ for which the accuracy
of the classifer $\vec{v}_i> \theta$
(or $\vec{v}_i< \theta$) was greatest -- where the
discrimination task was to distinguish triples that had one
word vs the other as their central word.
So for ``typical documentary footage'' and
``good documentary can'' the classifier should predict class
1 (``documentary''), for
``really special walk'' the classifier should predict class
2 (``special'').
Finally, of the 100 discrimination accuracies we chose the
largest one for this word pair.

On this discrimination task, SVD$^2$ was better than SVD$^1$ 55
times, the two were equal 15 times and SVD$^2$ was worse 30
times. 
On average, discrimination accuracy of SVD$^2$ was 0.7\% better
than that of SVD$^1$.
This is evidence that SVD$^2$ is
better for this discrimination task than SVD$^1$.

This indicates again that SVD$^2$ representations are
more focal
than SVD$^1$ representations: each dimension is
more likely to provide crucial information by itself as
opposed to only being useful in conjunction with other
dimensions. 

To illustrate in more detail why  this discrimination task is related to
focality, assume that for a particular 100-dimensional
representation $r$ of trigrams $t$,
the decision rule ``if $r(t)_{27}>0.2$ then `documentary' else
`special' '' (i.e., if the value of
dimension 27 is greater than 0.2, then the trigram center is
predicted to be ``documentary'', else ``special'') has an
accuracy of 0.99;
and that the decision rule ``if $r(t)_{27}>0.2$ and 
$r(t)_{91}<-0.1$
then `documentary' else
`special' '' 
has an
accuracy of 1.0. 
Then $M_{0.99}(f,\mbox{documentary-vs-special})=1$,
$M_{1.00}(f,\mbox{documentary-vs-special})=2$ and we can
view the representation $r$ as highly focal since a single
dimension suffices for high accuracy and two dimensions
achieve perfect classification results.

\section{1LAYER vs.\ 2LAYER}
We compare two
representations of a word trigram: (i) the 1LAYER
representation from
Section 1
and (ii) a 2LAYER representation that goes through
two rounds of autoencoding, which is a deep learning representation
in the sense that  layer 2 represents more
general and higher-level properties of the input than
layer 1.

\begin{figure}[tp]
\includegraphics[width=7cm]{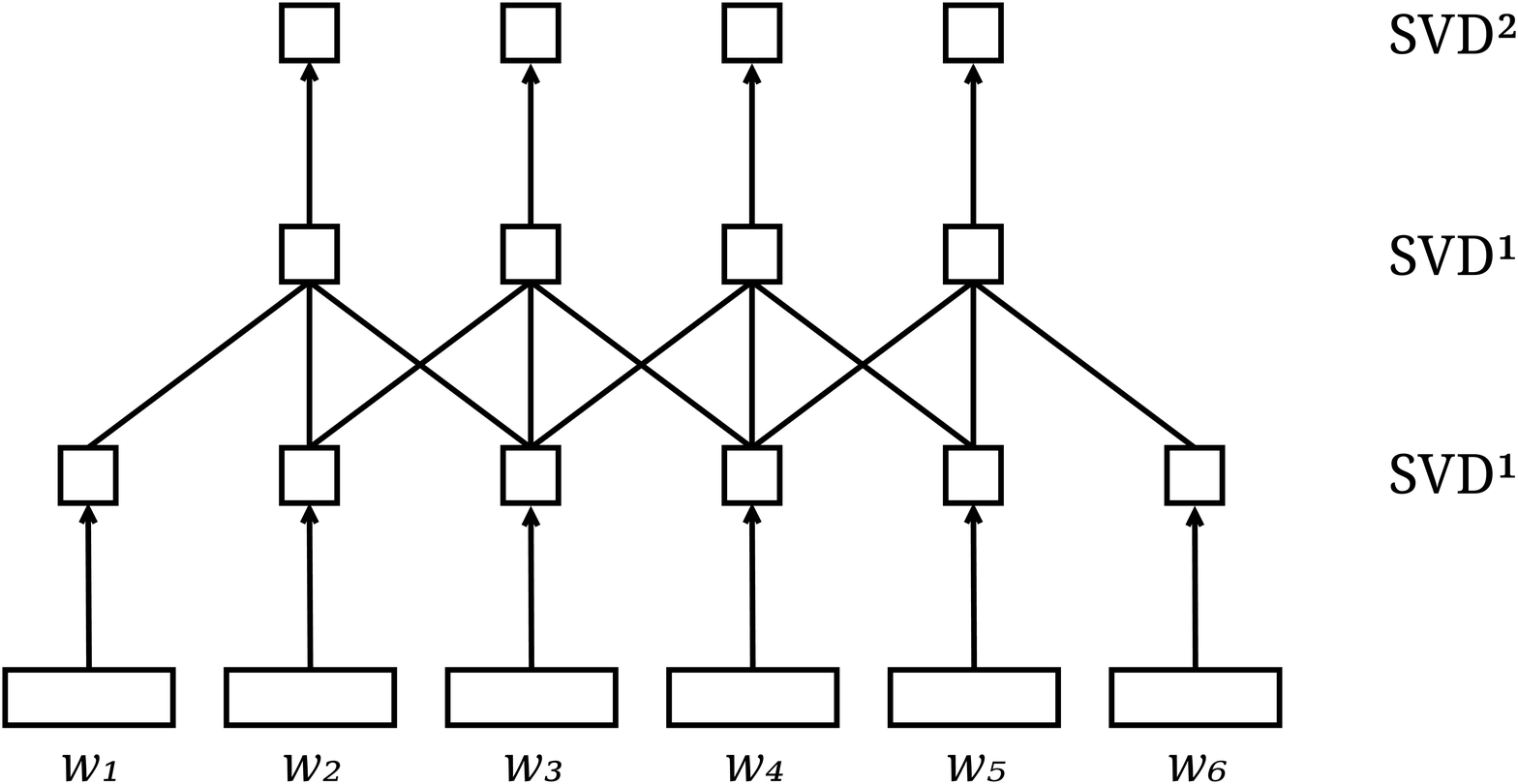}
    \caption{2LAYER architecture. Distributional word
      vectors form the bottom layer. 
Each word vector
is
      transformed by  SVD into a 100-dimensional
      vector (first layer ``SVD$^1$''). This layer
      constitutes the  1LAYER part of this architecture. 
A triple of vectors of three consecutive words 
is
      transformed by  SVD into a 100-dimensional
      vector (second layer ``SVD$^1$''). These vectors are then
      rotated (layer ``SVD$^2$''). The last two layers
      constitute the 2LAYER part of this architecture.}
\figlabel{layer2}
\end{figure}

The architecture of the 2LAYER  is depicted in \figref{layer2}.

To create 2LAYER representations,
we first create a vector for each of the 20701 word types
occurring in the
corpus. This vector is the concatenation of its left vector
and its right vector. The resulting 20701 $\times$ 500 matrix
is the input representation to SVD$^1$.
We again set $k=100$. A trigram is then represented as the
concatenation of three of these 100-dimensional vectors.
We apply
the SVD$^2$ construction algorithm to the resulting 100000 $\times$ 300
matrix and truncate to $k=100$.

We now have -- for each trigram -- two 
SVD$^2$ representations, the 1LAYER representation from
Section 1 and the 2LAYER representation we just described.
We compare these two trigram representations, again using
 the  task from Section 1: discrimination of the 100 pairs
 of words.

2LAYER is better than 1LAYER 64 times on this task, the same in 18
cases and worse in 18 cases. This is 
statistically significant ($p<.01$, binomial test) evidence that 2LAYER 
SVD$^2$ representations are more focal than 
1LAYER SVD$^2$ representations.

\section{Discussion}
\subsection{Focality}

One advantage of focal representations is that many
classifiers cannot handle conjunctions of several
features unless they are explicitly defined as separate features. Compare two
representations $\vec{x}$ and $\vec{x}'$ 
where $\vec{x}'$ is a rotation of $\vec{x}$ (as it
might be obtained by an SVD). Since one vector is a rotation
of the other, they contain exactly the same
information. However, if (i) an individual ``hidden unit''
of the rotated vector $\vec{x}'$ can directly be interpreted
as ``is verb'' (or a similar property like ``is adjective''
or ``takes NP argument'') and (ii) the same feature requires
a conjunction of several hidden units for $\vec{x}$, then
the rotated representation is superior for many upstream
statistical classifiers.

Focal representations can be argued to be closer to
biological reality than broadly distributed representations
\cite{thorpe10grandmother}; and they
have the nice property that
they become categorical in the limit. Thus, they include
categorical representations as a special case.

A final advantage of focal representations is that in
some convolutional architectures the input to the top-layer
statistical classifier consists
of maxima over HU (hidden unit) activations. E.g.,
one way to classify a sentence as having positive/negative
sentiment is to slide a neural network whose input is a window of $k$ words (e.g., $k=4$)
over it and to represent each window of $k$ words as
a vector of HU activations produced by the network. In a focal representation, the hidden units are
more likely to have clear semantics like ``the window
contains a positive sentiment word''.
In this type of scenario, taking the
maximum of activations over the $n-k+1$ sliding windows of a
sentence of length $n$ results in hidden units
with interpretable semantics like ``the activation
of the positive-sentiment HU of the window with the
highest activation for this HU''.
 These maximum
values are then a good basis for sentiment
classification of the sentence as a whole.

The notion of focality is similar to disentanglement
\cite{glorot11domain} -- in fact, the two notions may be
identical. However, 
\newcite{glorot11domain} introduce disentanglement in the
context of domain adaptation, focusing on the idea that
``disentangled'' hidden units capture general cross-domain
properties and for that reason are a good basis for domain
adaptation. The contributions of this paper are: 
proposing a way of measuring ``entanglement'' (i.e.,
measuring it as correlation),
defining
focality in terms of classification accuracy (a definition
that covers single hidden units as well as groups of hidden
units) and discussing the relationship to convolution and
biological systems.

It is important to point out that we have not addressed
how focality would be computed efficiently in  a particular
context. In theory, we could use brute force methods, but
these would be exponential in the number of dimensions
(systematic search over all subsets of dimensions). However,
certain interesting questions about focality can be answered
efficiently; e.g., if we have $M(f,R)=1$ for one
representation and $M(f,R')>1$ for another, then this can be
shown efficiently and in this case we have established
that $R$ is more focal than $R'$.

\subsection{mSVD method}
In this section, we will use the abbreviation \emph{mSVD} to
refer to a stacked applications of our method with an
arbitrary number of layers even though we only
experiment with $m=2$ in this paper (2LAYER,
2-layer-stacking).

SVD and other least squares methods are probably the 
most widely used dimensionality reduction techniques for the
type of matrices in natural language processing that we work with in this paper 
(cf.\ \cite{turney10vectorspace}). 
Stacking a second least squares method on
top of the first has not been considered widely because these types
of representations are usually used directly in vector
classifiers such as Rocchio and SVM (however, see the
discussion of
\cite{chen12marginalized} below). For this type of
classifier, performing a rotation has no effect on
classification performance. In contrast, our interest is to
use SVD$^2$ representations as part of a multilevel deep
learning architecture where the hidden unit
representations of any given layer are not simply interpreted
as a vector, but decisions of higher layers can be based on
individual dimensions.

The potential drawback of SVD and other least squares 
dimensionality reductions
is that they are linear: reduced
dimensions are linear combinations of orginal dimensions. To
overcome this limitation many nonlinear methods have been introduced:
probabilistic latent semantic indexing \cite{hofmann99plsi},
kernel principal component analysis
\cite{scholkopf98nonlinear}, matrix factorization techniques
that obey additional constraints -- such as non-negativity in
the case of
non-negative matrix factorization
\cite{lee99learning} -- ,
latent dirichlet allocation
\cite{blei03latent} and different forms of
autoencoding \cite{bengio09deep,chen12marginalized}. All of these can be viewed
as dimension reduction techniques that do not make the
simplistic assumptions of SVD and should therefore be able
to produce better representation if these simplistic 
assumptions are not appropriate for the domain in question.

However, this argument does not apply to the mSVD
method we propose in this paper since it is also
nonlinear. What should be investigated in the future is to
what extent the type of nonlinearity implemented by
mSVD offers advantages over other forms of nonlinear
dimensionality reduction; e.g., if the quality of the final
representations is comparable, then mSVD would have the
advantage of being more
efficient.

Finally, there is one big difference between
mSVD and deep learning representations such as those proposed by 
\newcite{hinton06deep},
\newcite{collobert2008unified} and
\newcite{socher12semantic}. Most deep learning
representations are induced in a setting that also includes
elements of supervised learning as is the case in contrastive
divergence or when labeled data are available for adjusting
initial representations produced by a process like
autoencoding or dimensionality reduction. 

This is the most
important open question related to the research presented
here: how can one modify hidden layer representations
initialized by multiple SVDs in a meaningful way?

The work most closely related to what we are proposing
is probably mDA
\cite{chen12marginalized} -- an approach we only became aware of
after the initial publication of this paper.
There are a number of differences between mDA and
mSVD. Non-linearity in mDA is achieved by classical deep
learning encoding functions like tanh() whereas we
renormalize vectors and then rotate them. Second, we do not
add noise to the input vectors -- mSVD is more efficient for
this reasons, but it remains to be seen if it can achieve
the same level of performance as mDA. Third, the mSVD 
architecture proposed here, which changes the objects to be
represented from small frequent units to larger less
frequent units when going one layer up, can be seen as an
alternative (though less general since it's customized for
natural language) way of extending to very high dimensions.

\section{Conclusion}

As a next step 
a direct comparison should be performed of
SVD$^2$ with traditional deep learning
\cite{hinton06deep}. As we have argued,
SVD$^2$ would be an interesting alternative to
deep learning initialization methods currently used 
since SVD is efficient and
a simple and well understood formalism. But this argument is
only valid if the resulting representations are of
comparable quality. Datasets and tasks for this comparative
evaluation could e.g. be those of 
\newcite{turian10word},
\newcite{maas11learning}, and
\newcite{socher11recursive}.

\bibliographystyle{coling}
\bibliography{buecher}

\begin{thebibliography}{}

\bibitem[\protect\citename{Anandkumar \bgroup et al.\egroup
  }2012]{anandkumar12two}
Anandkumar, Animashree, Dean~P. Foster, Daniel Hsu, Sham~M. Kakade, and Yi-Kai
  Liu.
\newblock 2012.
\newblock Two svds suffice: Spectral decompositions for probabilistic topic
  modeling and latent dirichlet allocation.
\newblock {\em CoRR}, abs/1204.6703.

\bibitem[\protect\citename{Bengio}2009]{bengio09deep}
Bengio, Yoshua.
\newblock 2009.
\newblock Learning deep architectures for {AI}.
\newblock {\em Foundations and Trends in Machine Learning}, 2(1):1--127.

\bibitem[\protect\citename{Blei \bgroup et al.\egroup }2003]{blei03latent}
Blei, David~M., Andrew~Y. Ng, and Michael~I. Jordan.
\newblock 2003.
\newblock Latent {D}irichlet allocation.
\newblock {\em JMLR}, 3:993--1022.

\bibitem[\protect\citename{Chen \bgroup et al.\egroup
  }2012]{chen12marginalized}
Chen, Minmin, Zhixiang~Eddie Xu, Kilian~Q. Weinberger, and Fei Sha.
\newblock 2012.
\newblock Marginalized denoising autoencoders for domain adaptation.
\newblock In {\em ICML}.

\bibitem[\protect\citename{Collobert and Weston}2008]{collobert2008unified}
Collobert, Ronan and Jason Weston.
\newblock 2008.
\newblock A unified architecture for natural language processing: Deep neural
  networks with multitask learning.
\newblock In {\em ICML}.

\bibitem[\protect\citename{Glorot \bgroup et al.\egroup }2011]{glorot11domain}
Glorot, Xavier, Antoine Bordes, and Yoshua Bengio.
\newblock 2011.
\newblock Domain adaptation for large-scale sentiment classification: A deep
  learning approach.
\newblock In {\em ICML}, pages 513--520.

\bibitem[\protect\citename{Hinton \bgroup et al.\egroup }2006]{hinton06deep}
Hinton, Geoffrey~E., Simon Osindero, and Yee-Whye Teh.
\newblock 2006.
\newblock A fast learning algorithm for deep belief nets.
\newblock {\em Neural Computation}, 18(7):1527--1554.

\bibitem[\protect\citename{Hofmann}1999]{hofmann99plsi}
Hofmann, Thomas.
\newblock 1999.
\newblock Probabilistic latent semantic indexing.
\newblock In {\em SIGIR}, pages 50--57.

\bibitem[\protect\citename{Lee and Seung}1999]{lee99learning}
Lee, David~D. and H.~Sebastian Seung.
\newblock 1999.
\newblock Learning the parts of objects by non-negative matrix factorization.
\newblock {\em Nature}, 401:788.

\bibitem[\protect\citename{Maas \bgroup et al.\egroup }2011]{maas11learning}
Maas, Andrew~L., Raymond~E. Daly, Peter~T. Pham, Dan Huang, Andrew~Y. Ng, and
  Christopher Potts.
\newblock 2011.
\newblock Learning word vectors for sentiment analysis.
\newblock In {\em ACL}, pages 142--150.

\bibitem[\protect\citename{Pang and Lee}2004]{Pang+Lee:04a}
Pang, Bo and Lillian Lee.
\newblock 2004.
\newblock A sentimental education: Sentiment analysis using subjectivity
  summarization based on minimum cuts.
\newblock In {\em Proc.\ of ACL}.

\bibitem[\protect\citename{Sch{\"o}lkopf \bgroup et al.\egroup
  }1998]{scholkopf98nonlinear}
Sch{\"o}lkopf, Bernhard, Alex~J. Smola, and Klaus-Robert M{\"u}ller.
\newblock 1998.
\newblock Nonlinear component analysis as a kernel eigenvalue problem.
\newblock {\em Neural Computation}, 10(5):1299--1319.

\bibitem[\protect\citename{Sch{\"u}tze}1995]{schutze95distributional}
Sch{\"u}tze, Hinrich.
\newblock 1995.
\newblock Distributional part-of-speech tagging.
\newblock In {\em Conference of the European Chapter of the Association for
  Computational Linguistics}, pages 141--148.

\bibitem[\protect\citename{Socher \bgroup et al.\egroup
  }2011]{socher11recursive}
Socher, Richard, Jeffrey Pennington, Eric~H. Huang, Andrew~Y. Ng, and
  Christopher~D. Manning.
\newblock 2011.
\newblock Semi-supervised recursive autoencoders for predicting sentiment
  distributions.
\newblock In {\em EMNLP}, pages 151--161.

\bibitem[\protect\citename{Socher \bgroup et al.\egroup
  }2012]{socher12semantic}
Socher, Richard, Brody Huval, Christopher~D. Manning, and Andrew~Y. Ng.
\newblock 2012.
\newblock Semantic compositionality through recursive matrix-vector spaces.
\newblock In {\em EMNLP-CoNLL}, pages 1201--1211.

\bibitem[\protect\citename{Thorpe}2010]{thorpe10grandmother}
Thorpe, Simon.
\newblock 2010.
\newblock Grandmother cells and distributed representations.
\newblock In Kriegeskorte, Nikolaus and Gabriel Kreiman, editors, {\em
  Understanding visual population codes. Toward a common multivariate framework
  for cell recording and functional imaging}. MIT Press.

\bibitem[\protect\citename{Turian \bgroup et al.\egroup }2010]{turian10word}
Turian, Joseph, Lev-Arie Ratinov, and Yoshua Bengio.
\newblock 2010.
\newblock Word representations: A simple and general method for semi-supervised
  learning.
\newblock In {\em ACL}, pages 384--394.

\bibitem[\protect\citename{Turney and Pantel}2010]{turney10vectorspace}
Turney, Peter~D. and Patrick Pantel.
\newblock 2010.
\newblock From frequency to meaning: Vector space models of semantics.
\newblock {\em J. Artif. Intell. Res. (JAIR)}, 37:141--188.

\end{thebibliography}

\end{document}